\documentclass[10pt,twocolumn,letterpaper]{article}

\usepackage{wacv}              % To produce the CAMERA-READY version
\definecolor{wacvblue}{rgb}{0.21,0.49,0.74}
\usepackage[pagebackref,breaklinks,colorlinks,allcolors=wacvblue]{hyperref}

\def\wacvPaperID{1970} % *** Enter the WACV Paper ID here
\def\confName{WACV}
\def\confYear{2026}

\usepackage{booktabs} % 문서 preamble에 추가
\usepackage{placeins}  % \FloatBarrier 사용 가능하게 함
\usepackage{multirow}
\usepackage{microtype}  % 미세 조정 자동화 (preamble에서 한 번만)

\usepackage{amssymb}     % \mathbb{1} 등 기호 사용을 위한 패키지
\usepackage{amsmath}     % 수식 정렬 등 수학 용도
\usepackage{pifont}
\usepackage{dsfont}
\usepackage{array}

\usepackage{color}

\title{NOVO: Bridging LLaVA and SAM \\with Visual-only Prompts for Reasoning Segmentation}

\author{
Kyung-Yoon Yoon and Yeong-Jun Cho\textsuperscript{*}\\
Department of Artificial Intelligence Convergence\\
Chonnam National University, Gwangju 61186, South Korea\\
{\tt\small \{kyungyoon201, yj.cho\}@jnu.ac.kr}
}

\begin{document}
\maketitle

\begin{abstract}
% 우리는 Reasoning Segmentation via Visual Prompts (RSVP)라는 새로운 프레임워크를 제안한다. 이 프레임워크는 \textit{시각 프롬프트(visual-only prompts)}를 통해 비전-언어 모델(VLM)과 세분화 모델 간의 연결을 가능하게 한다.
% 기존의 방법들이 텍스트 기반의 \texttt{<SEG>} 임베딩을 세분화 모델에 직접 주입하는 것과 달리, RSVP는 이를 Segment Anything Model (SAM)과 호환되는 coarse mask와 point prompt로 변환함으로써, SAM의 사전 학습 구조와의 정렬을 유지한다.
% 또한, RSVP는 추가 학습 없이 동작하는 refinement 모듈을 도입하여 시각적 오류를 줄이고 세분화 마스크의 품질을 향상시켜 경계 정확도를 높이고 인스턴스 수준의 세분화를 가능하게 한다.
% 더불어, 우리는 총 918개의 이미지, 2,533개의 인스턴스 마스크, 다양한 reasoning 질의로 구성된 대규모 벤치마크 데이터셋 \textbf{RISeg}도 새롭게 제안한다.
% 실험 결과, RSVP는 다양한 평가 지표 및 모델 규모 전반에서 state-of-the-art 성능을 달성하였으며, reasoning segmentation에서의 효과성과 확장 가능성을 입증하였다.
% In this study, we propose Reasoning Segmentation via Visual Prompts (RSVP), a novel framework that bridges vision-language models (VLMs) and segmentation models through visual-only prompts.
In this study, we propose NOVO (NO text, Visual-Only prompts), a novel framework that bridges vision-language models (VLMs) and segmentation models through visual-only prompts.
Unlike prior approaches that feed text-derived \texttt{<SEG>} token embeddings into segmentation models, NOVO instead generates a coarse mask and point prompts from the VLM output.
These visual prompts are compatible with the Segment Anything Model (SAM), preserving alignment with its pretrained capabilities.
To further enhance boundary quality and enable instance-level segmentation, we introduce a training-free refinement module that reduces visual artifacts and improves the quality of segmentation masks.
We also present \texttt{RISeg}, a new benchmark comprising 918 images, 2,533 instance-level masks, and diverse reasoning queries to evaluate this task.
Experiments demonstrate that NOVO achieves state-of-the-art performance across multiple metrics and model sizes, demonstrating its effectiveness and scalability in reasoning segmentation.

\noindent\textbf{Code and Dataset:} \url{https://WILL_BE_SOON}
\end{abstract}

%%%%%%%%% BODY TEXT
\section{Introduction}
\label{sec:1}

\begin{figure}[t]
    \hfill
    \includegraphics[width=\columnwidth]{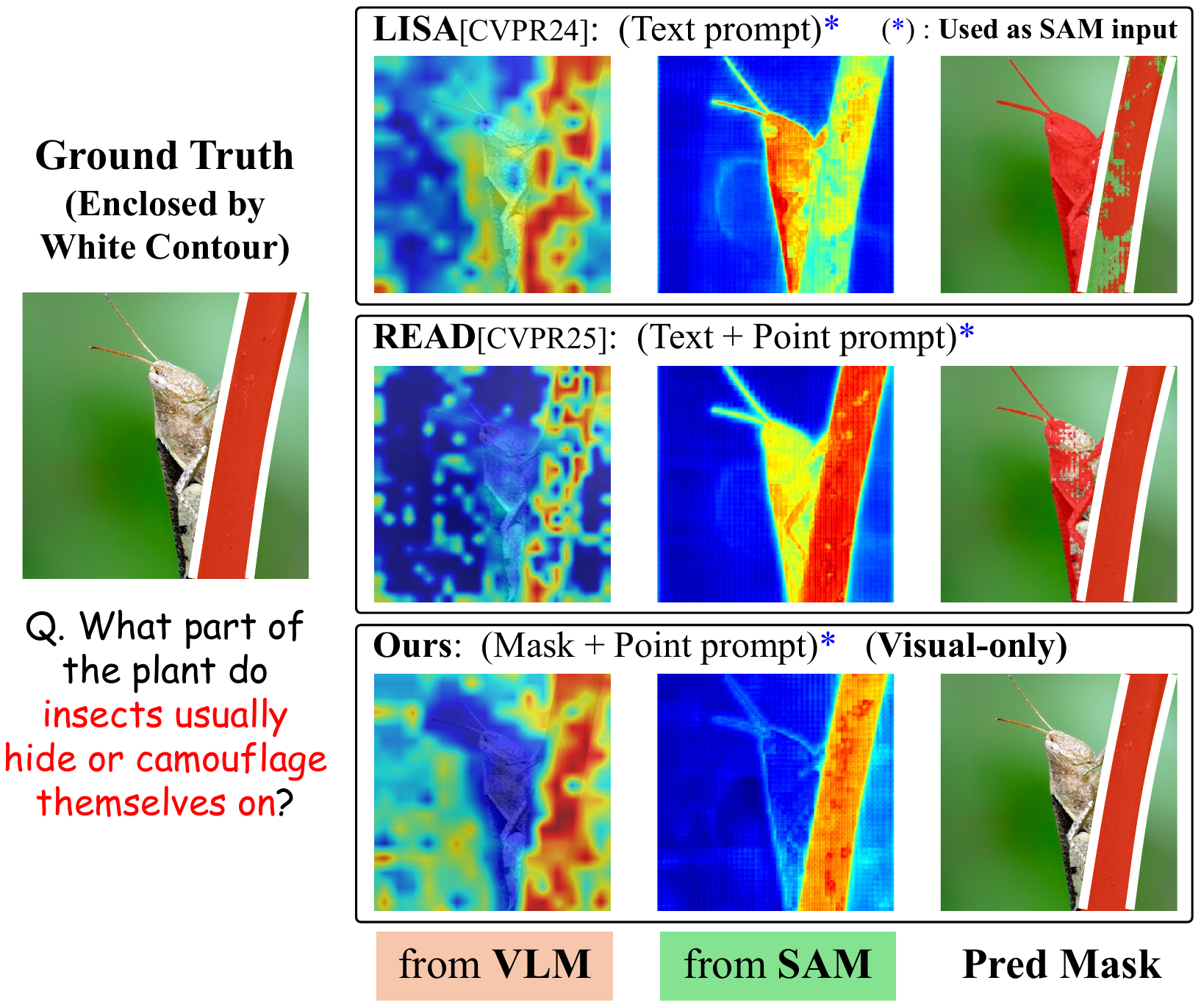}
    \caption{
    Comparison of VLM activation maps, SAM mask logits, and predicted masks from different models.
    Our method, which uses visual prompts only, produces masks that better align with the semantics of the input query. In contrast, LISA and READ often show a mismatch between VLM activations and SAM mask logits.
    Best viewed in color.
    }
    \label{fig:first_figure}
\end{figure}

In real-world scenarios, segmentation is not always about detecting clearly labeled objects.
It often requires understanding visual context and reasoning over indirect descriptions.
% For example, Fig.\ref{fig:first_figure} shows a case where the goal is to segment the region described by the prompt, ``find the part of the image where the car is reflected,'' rather than the car itself.
For example, Fig.\ref{fig:first_figure} shows a case where the goal is to segment the region described by the prompt, ``find the part of the plant where insects usually hide or camouflage themselves,'' rather than directly segmenting the leaves themselves.
This challenge is known as Reasoning Segmentation, which deals with complex, query-based segmentation.
It takes an implicit text query as input and predicts a binary mask for the relevant region.
Unlike traditional Referring Segmentation~\cite{kazemzadeh2014referitgame, nagaraja2016modeling}, which targets explicitly mentioned objects, this task requires deeper language understanding and visual reasoning.

% A common strategy in reasoning segmentation is to extract the \texttt{<SEG>} token embedding from a vision-language model (VLM) and use it as a text prompt for segmentation.
A common strategy in reasoning segmentation is to extract the \texttt{<SEG>} token embedding from a vision-language model (VLM) and use it as a prompt for segmentation.
Methods like LISA~\cite{lai2024lisa} and READ~\cite{qian2025reasoning} follow this approach using SAM~\cite{kirillov2023segment} for segmentation.
However, SAM is trained to respond to visual prompts--such as points, boxes, and masks--rather than text embeddings.
Therefore, directly using a text-based \texttt{<SEG>} token embedding may not align with this training, potentially leading to inaccurate predictions.
% To examine this issue, we compare VLM attention maps and SAM’s predicted masks using existing models (Fig.~\ref{fig:first_figure}).
To examine this issue, we compare VLM activation maps and SAM’s predicted masks using existing models, as shown in Fig.~\ref{fig:first_figure}.
While the VLM activation in existing methods (e.g., LISA, READ) often points to the correct region, the corresponding SAM output masks frequently fail to align with it.
This motivates a new approach that transforms the output of VLMs~(i.e., \texttt{<SEG>} token embedding) into visual prompts, allowing SAM to accurately segment regions based on the input query.

% To address these limitations, we propose a new framework called Reasoning Segmentation via Visual Prompt (RSVP) as described in Sec.~\ref{sec:4}. 
To address these limitations, we propose a new framework called NOVO (NO text, Visual-Only prompts) for reasoning segmentation as described in Sec.~\ref{sec:4}.
Instead of directly feeding text-based embeddings into the segmentation model, NOVO performs reasoning segmentation using only visual prompts.
Specifically, NOVO generates two types of visual prompts: a coarse mask prompt derived from the VLM activation map, and point prompts sampled from regions with high semantic relevance.
These visual prompts are then used to guide SAM, without relying on any text-based inputs.

To further boost performance, we propose a refinement module (Sec.~\ref{sec:4.5}) that selects the most relevant masks from SAM’s outputs.
This design also allows NOVO to fully leverage SAM’s segmentation capabilities, resulting in accurate and semantically aligned outputs.
Interestingly, our refinement process naturally extends to instance-level reasoning segmentation, without requiring additional model modifications.
To support evaluation of instance-level reasoning segmentation, we construct a new benchmark dataset, \texttt{RISeg} (Sec.~\ref{sec:5}).
It includes 918 images from COCO, 2,533 instance masks, and diverse reasoning queries automatically generated using GPT-4o and verified by human experts.

NOVO outperforms previous state-of-the-art methods and achieves new best results on the \texttt{ReasonSeg} benchmark.
Specifically, NOVO achieves 62.4\% gIoU and 65.7\% cIoU on the test set when using a 7B vision-language backbone, outperforming the previous best by +3.9\% and +6.2\%, respectively.
With a stronger 13B backbone, NOVO further improves to 65.3\% gIoU and 66.0\% cIoU, establishing a new state-of-the-art.
In addition, our NOVO Refinement significantly improves boundary quality by mitigating visual artifacts such as holes and imprecise contours.
It also enables instance-level segmentation without additional training and consistently outperforms prior methods on the \texttt{RISeg} benchmark.

Our main contributions are summarized as follows:
\begin{itemize}
    % \item We conduct an analytical investigation into the limitation of existing reasoning segmentation models, where the VLM correctly identifies the target object, yet the segmentation output from SAM does not align with the actual object location.
    \item We propose a novel reasoning segmentation framework, NOVO, that uses only visual prompts to better align the VLM’s understanding with the segmentation output.
    \item We introduce a refinement module that enhances mask quality and enables instance-level segmentation without requiring additional training.
    \item We construct a new benchmark dataset, \texttt{RISeg}, designed to evaluate instance-level reasoning segmentation.
    \item Extensive experiments show that NOVO consistently outperforms previous methods and achieves new state-of-the-art results on reasoning segmentation benchmarks.
\end{itemize}
% NOVO is the first to show that a purely visual prompt pipeline can outperform text-driven approaches, highlighting the need to match tasks with the pretraining modality.

\section{Related Works}
\label{sec:2}

% \paragraph{Prompt-based Segmentation.}
\noindent\textbf{Prompt-based Segmentation.} \quad
Recent advances in segmentation have shifted from fixed class-based pixel prediction to prompt-based methods that adapt flexibly to user inputs.
The Segment Anything Model (SAM)~\cite{kirillov2023segment} is a foundation model that enables high-quality segmentation using various user prompts (e.g., text, points, boxes, masks).
Building on this, recent research has explored the use of natural language prompts in addition to visual ones.
Approaches such as OVSeg~\cite{liang2023open}, GRES~\cite{liu2023gres}, X-Decoder~\cite{zou2023generalized}, and SEEM~\cite{zou2023segment} introduce unified frameworks that leverage CLIP~\cite{radford2021learning} or other vision-language models to address diverse segmentation tasks.
These models aim to improve zero-shot inference while supporting both open-vocabulary and referring segmentation scenarios.
However, they still struggle with queries requiring contextual or implicit reasoning, leading to extensions toward models with stronger reasoning ability.

% \paragraph{Reasoning Segmentation.} 
% \noindent\textbf{Reasoning Segmentation.}
% To address queries requiring contextual or implicit understanding, the task of ``\textit{Reasoning Segmentation}'' has been introduced.
% This task aims to segment objects based on language queries that require high-level reasoning, such as indirect or context-dependent descriptions.
% LISA~\cite{lai2024lisa} first introduced the concept of reasoning segmentation by using the \texttt{<SEG>} token embedding from a vision-language model (VLM) as a prompt to guide a segmentation model.
% SESAME~\cite{wu2024see} extended this approach to handle queries involving non-existent objects (i.e., false premises).
% READ~\cite{qian2025reasoning} further analyzed the semantic function of the \texttt{<SEG>} token and incorporated highly activated points derived from similarity maps.
% LISA++~\cite{yang2023lisa++} built upon this framework by introducing multiple \texttt{<SEG>} tokens to support instance-level reasoning and interactive segmentation.
% These methods typically use SAM as the segmentation backbone, guided by the \texttt{<SEG>} token embedding from a vision-language model.
% Although SAM has been exposed some text prompts during training, it is primarily trained with visual prompts. As a result, using language-based embeddings with SAM may cause a mismatch with its training objectives and lead to reduced segmentation performance.
\noindent\textbf{Reasoning Segmentation.} \quad
To address queries requiring contextual or implicit understanding, the task of ``\textit{Reasoning Segmentation}'' has been introduced.
This task aims to segment objects based on language queries that require high-level reasoning, such as indirect or context-dependent descriptions.
LISA~\cite{lai2024lisa} first introduced the concept of reasoning segmentation by using the \texttt{<SEG>} token embedding from a vision-language model (VLM) as a prompt to guide a segmentation model.
SESAME~\cite{wu2024see} extended this approach to handle queries involving non-existent objects (i.e., false premises).
READ~\cite{qian2025reasoning} further analyzed the semantic function of the \texttt{<SEG>} token and incorporated highly activated points derived from similarity maps.
LISA++~\cite{yang2023lisa++} built upon this framework by introducing multiple \texttt{<SEG>} tokens to support instance-level reasoning and interactive segmentation.
% \tp{Subsequent works such as GSVA, SAM4MLLM, PixelLM, GLaMM, and POPEN further expand reasoning segmentation by addressing absent objects, introducing SAM-friendly prompts, adopting end-to-end decoding, or applying preference-based optimization~\cite{xia2024gsva,chen2024sam4mllm,ren2024pixellm,rasheed2024glamm,zhu2025popen}.
These methods generally use SAM as the segmentation backbone, which is mainly trained with visual prompts.
As a result, using language-based embeddings with SAM may cause a mismatch with its training objectives and lead to reduced segmentation performance.
Subsequent works such as GSVA~\cite{xia2024gsva}, SAM4MLLM~\cite{chen2024sam4mllm}, GLaMM~\cite{rasheed2024glamm}, POPEN~\cite{zhu2025popen}, and RSVP~\cite{lu2025rsvp} further expand reasoning segmentation by handling absent objects, MLLM-guided prompts, adopting end-to-end decoding, applying preference optimization, or leveraging chain-of-thought.

% \paragraph{Vision-Language Models (VLMs).} 
\noindent\textbf{Vision-Language Models (VLMs).} \quad
Recent advances in vision-language models (VLMs) have significantly influenced the development of text-based reasoning segmentation.
Representative multimodal models include Flamingo~\cite{alayrac2022flamingo}, BLIP-2~\cite{li2023blip}, mPLUG-Owl~\cite{ye2023mplug}, OTTER~\cite{li2025otter}, LLaVA~\cite{liu2023visual}, and MiniGPT-4~\cite{zhu2023minigpt}.
These models have shown strong performance on complex tasks such as instruction-based question answering, explanation generation, and visual reasoning.
% These models align visual and language features, enabling contextual understanding and user intent recognition, which are essential capabilities for reasoning segmentation.
These models integrate visual and language features, enabling contextual understanding and user intent recognition, which are essential capabilities for reasoning segmentation.

\begin{figure*} [t]
    \centering
    \includegraphics[width=0.95\textwidth]{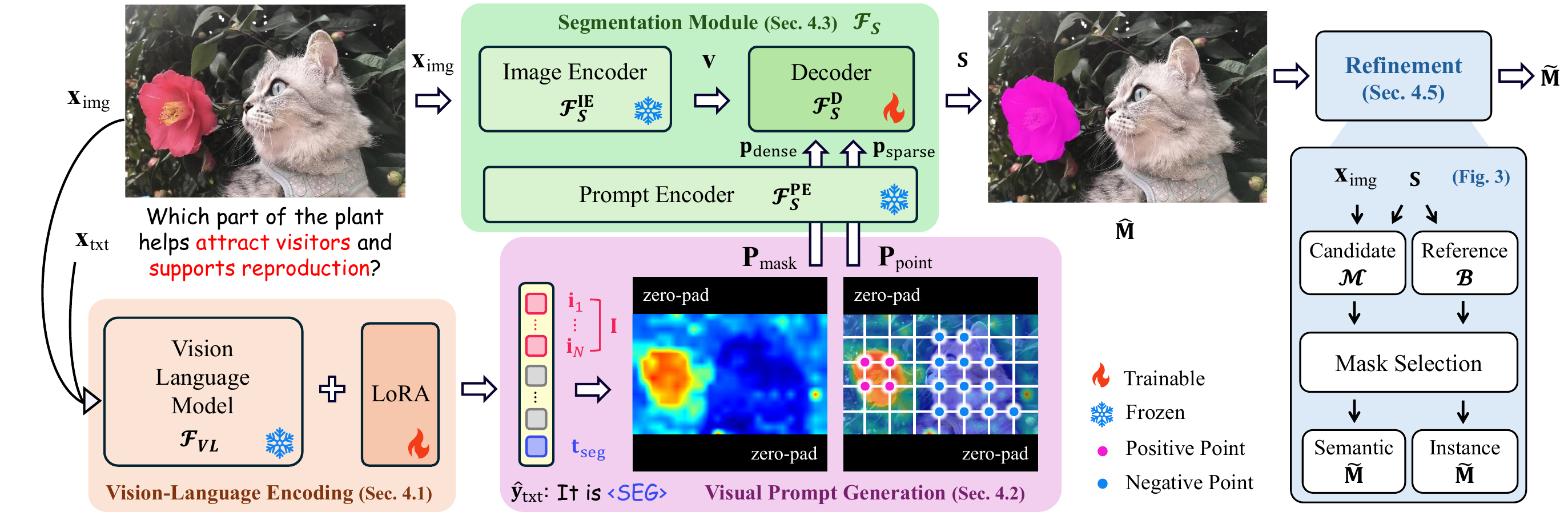}
    \caption{
    Overview of the proposed NOVO. It encodes an input image $\mathbf{x}_{\text{img}}$ and a reasoning text query $\mathbf{x}_{\text{txt}}$ via a VLM to extract the \texttt{<SEG>} token embedding $\mathbf{t}_{\text{seg}}$.
    Together with image patch embeddings $\mathbf{I}$, it generates the mask prompt $\mathbf{P}_{\text{mask}}$ and point prompt $\mathbf{P}_{\text{point}}$,
    which are passed as the sole inputs to the NOVO's segmentation module to produce a segmentation mask $\hat{\mathbf{M}}$.
    The predicted mask can be refined using our proposed method in Sec.~\ref{sec:4.5}, which not only improves segmentation quality but also enables instance-level mask generation.}
    \label{fig:main_figure}
    \vspace{-3pt}
\end{figure*}

\section{Motivation and Main Idea}
\label{sec:3}
Reasoning segmentation aims to generate a binary mask $\hat{\mathbf{M}} \in \{0, 1\}^{h \times w}$ from a text query $\mathbf{x}_{\text{txt}}$ and an input image $\mathbf{x}_{\text{img}}$, defined as 
$\hat{\mathbf{M}} = \mathcal{F}_{\theta}(\mathbf{x}_{\text{img}}, \mathbf{x}_{\text{txt}}).$
The model \(\mathcal{F}_{\theta}\), composed of a vision-language model (VLM) $\mathcal{F}_\text{VL}$ and a segmentation model $\mathcal{F}_\text{S}$, is trained to optimize the parameters $\theta$ such that the predicted mask $\hat{\mathbf{M}}$ closely matches the ground-truth mask.
For the segmentation model, previous studies~\cite{lai2024lisa,qian2025reasoning} adopted Segment Anything Model (SAM)~\cite{kirillov2023segment}.
The VLM infers the location of the object described by the text query, while the SAM generates a precise segmentation mask based on that location.

As shown in Fig.~\ref{fig:first_figure}, we investigated the causes of performance degradation in reasoning segmentation.  
While the VLM often accurately infers the target region from the text query, existing methods such as LISA~\cite{lai2024lisa} and READ~\cite{qian2025reasoning} often fail to leverage this information.  
Moreover, the inconsistency between the outputs of the VLM and SAM leads to inaccurate reasoning masks.
We note that SAM has been extensively pretrained on large-scale data using visual prompts such as points and boxes rather than text-based prompts.  
However, relying on text-based prompts in previous methods (e.g., LISA and READ) may not sufficiently activate the pretrained segmentation capabilities of SAM.

% Rather than enhancing SAM's understanding of text prompts through additional training,
% we propose a novel approach called Reasoning Segmentation via Visual Prompts (RSVP), which transforms VLM outputs into visual prompts that align with SAM’s pretraining format.
Rather than enhancing SAM's understanding of text prompts through additional training,
we propose a novel approach called NOVO (NO text, Visual-Only prompts), which transforms VLM outputs into visual prompts that align with SAM’s pretraining format.
This design more effectively leverages SAM’s pretrained capabilities for reasoning segmentation.
To further enhance segmentation quality, we also propose NOVO Refinement, a training-free method that converts segmentation logits into point prompts and uses SAM to generate precise instance-level masks.

\section{Proposed NOVO}
\label{sec:4}
% We propose Reasoning Segmentation via Visual Prompts (RSVP) as illustrated in Fig.~\ref{fig:main_figure}.
We propose NOVO (NO text, Visual-Only prompts) for reasoning segmentation as illustrated in Fig.~\ref{fig:main_figure}.
It consists of four main stages: (1) Vision-Language Encoding, (2) Visual Prompt Generation, (3) Segmentation Module, and (4) NOVO Refinement.

\subsection{Vision-Language Encoding}

\label{sec:4.1}
In NOVO, we utilize LLaVA~\cite{liu2023visual} as the vision-language model~(VLM), which takes an image $\mathbf{x}_{\text{img}}$ and a reasoning-based text query $\mathbf{x}_{\text{txt}}$ as input. 
The model encodes the target region for segmentation using the \texttt{<SEG>} token along with image embeddings and the corresponding token embedding.
LLaVA’s image encoder requires a fixed square aspect ratio, and while some prior approaches adopt a center-cropping strategy, we instead apply zero-padding based on the longer side to preserve the original visual information.
For consistency, we apply the same padding strategy to the segmentation model described in Sec.~\ref{sec:4.3}.

An input image ${\mathbf{x}}_{\text{img}}$ is transformed into $N$ patch embeddings via CLIP's image encoder in LLaVA and projected to match the dimensionality of the LLM’s hidden states.  
These image embeddings are then concatenated with the token embeddings of the input query $\mathbf{x}_{\text{txt}}$, forming a unified multimodal sequence that is processed by the LLM.
As a result, the model generates an output text response $\mathbf{\hat{y}}_\text{txt}$, which includes a \texttt{<SEG>} token.
Although $\hat{\mathbf{y}}_\text{txt}$ is the text output of the VLM, we additionally extract two components from its final hidden states:
the image patch embeddings $\mathbf{I} = [\mathbf{i}_1, \ldots, \mathbf{i}_N] \in \mathbb{R}^{N \times d}$ and the \texttt{<SEG>} token embedding $\mathbf{t}_{\text{seg}} \in \mathbb{R}^d$, where $d$ is the embedding dimension.
The model is trained to ensure that $\mathbf{t}_{\text{seg}}$ effectively encodes the semantics of the region to be segmented.

\subsection{Visual Prompt Generation}
\label{sec:4.2}
We explain how to transform the VLM’s output embeddings into visual prompts suitable for SAM input.  
Inspired by READ~\cite{qian2025reasoning}, we use the image embeddings $\mathbf{I}$ and the \texttt{<SEG>} token embedding $\mathbf{t}_{\text{seg}}$ to generate the mask prompt $\mathbf{P}_{\text{mask}}$ and the point prompt $\mathbf{P}_{\text{point}}$.

First, the mask prompt is computed by measuring the cosine similarity between $\mathbf{t}_{\text{seg}}$ and each image patch embedding $\mathbf{i}_n$, resulting in a coarse activation map that highlights the region indicated by the text query $\mathbf{x}_{\text{txt}}$ as follows:
{\small
    \begin{equation}
    \mathbf{P}_{\text{mask}} = \left[\text{sim}(\mathbf{t}_{\text{seg}}, \mathbf{i}_1), \text{sim}(\mathbf{t}_{\text{seg}}, \mathbf{i}_2),\ldots,\text{sim}(\mathbf{t}_{\text{seg}}, \mathbf{i}_N)\right],
    \label{eq:mask_prompt_en}
    \end{equation}
}
where $\text{sim}(\mathbf{a}, \mathbf{b}) = \frac{\mathbf{a} \cdot \mathbf{b}}{\|\mathbf{a}\| \|\mathbf{b}\|}$ denotes the cosine similarity function.  
Initially, $\mathbf{P}_{\text{mask}}$ is an $N$-dimensional vector, which is then reshaped and upsampled via bilinear interpolation to form a $256 \times 256$ visual prompt map.

Additionally, based on the high and low activation values in \(\mathbf{P}_{\text{mask}}\), we extract the coordinates of the top and bottom-ranked positions to construct the point prompt \(\mathbf{P}_{\text{point}}\), consisting of positive and negative points. This helps emphasize semantically relevant and irrelevant regions, respectively.
Unlike previous reasoning segmentation methods such as LISA~\cite{lai2024lisa}, which use only text embeddings, and READ~\cite{qian2025reasoning}, which combines text embeddings with point prompts, our approach relies solely on visual inputs.
This simple yet focused design improves reasoning segmentation performance by leveraging only visual prompts, which align more directly with SAM’s capabilities.

\subsection{Segmentation Module}
\label{sec:4.3}
To generate accurate segmentation masks, we adopt the Segment Anything Model (SAM), a general-purpose segmentation model capable of handling various visual prompts. SAM consists of three major components: the image encoder $\mathcal{F}^{\text{IE}}_S$, the prompt encoder $\mathcal{F}^{\text{PE}}_S$, and the mask decoder $\mathcal{F}^{\text{D}}_S$.

First, the image encoder $\mathcal{F}^{\text{IE}}_S$ transforms the input image ${\mathbf{x}}_{\text{img}}$ into a high-dimensional image embedding, denoted as $
\mathbf{v} = \mathcal{F}^{\text{IE}}_S({\mathbf{x}}_{\text{img}})$.
The prompt encoder $\mathcal{F}^{\text{PE}}_S$ converts the visual prompts $\mathbf{P}_{\text{mask}}, \mathbf{P}_{\text{point}}$ into dense and sparse prompt embeddings respectively, as follows:
\begin{equation}
\mathbf{p}_{\text{dense}} = \mathcal{F}^{\text{PE}}_S(\mathbf{P}_{\text{mask}}), \quad
\mathbf{p}_{\text{sparse}} = \mathcal{F}^{\text{PE}}_S(\mathbf{P}_{\text{point}}),
\end{equation}
where $\mathcal{F}^{\text{PE}}_S$ is designed as a unified module that processes different prompt types, such as masks and points, by internally branching based on the input type.
Finally, the mask decoder $\mathcal{F}^{\text{D}}_S$ takes $\mathbf{v}$, $\mathbf{p}_{\text{dense}}$, and $\mathbf{p}_{\text{sparse}}$ as input, and outputs segmentation logits $\mathbf{S}$, which are then binarized to produce the binary segmentation mask:
\begin{equation}
\hat{\mathbf{M}} = \mathcal{F}^{\text{D}}_S(\mathbf{v},\ \mathbf{p}_{\text{dense}},\ \mathbf{p}_{\text{sparse}}).
\end{equation}
This architecture allows the segmentation model in NOVO to fully exploit the cues provided by visual prompts.

\subsection{Loss Functions}
\label{sec:4.4}

To train networks, we adopt a multi-objective training scheme that jointly optimizes the text generation and mask prediction losses to improve both language reasoning and visual segmentation capabilities. The text generation loss $\mathcal{L}_{\text{txt}}$ encourages the VLM to generate natural and accurate responses to the given text query. It is formulated as a weighted cross-entropy loss between the generated response $\hat{\mathbf{y}}_{\text{txt}}$ and the ground-truth response $\mathbf{y}_{\text{txt}}$:
\begin{equation}
\mathcal{L}_{\text{txt}} = \lambda_{\text{CE}} \cdot \mathcal{L}_{\text{CE}}(\hat{\mathbf{y}}_{\text{txt}}, \mathbf{y}_{\text{txt}}).
\label{eq:txt_loss}
\end{equation}
The mask prediction loss $\mathcal{L}_{\text{mask}}$ encourages the predicted mask $\hat{\mathbf{M}}$, generated based on visual prompts, to align with the ground-truth mask $\mathbf{M}$.
We adopt a weighted combination of Binary Cross Entropy (BCE) and Dice loss, defined as:
\begin{equation}
\mathcal{L}_{\text{mask}} = \lambda_{\text{BCE}} \cdot \mathcal{L}_{\text{BCE}}(\hat{\mathbf{M}}, \mathbf{M}) + \lambda_{\text{Dice}} \cdot \mathcal{L}_{\text{Dice}}(\hat{\mathbf{M}}, \mathbf{M}),
\label{eq:mask_loss}
\end{equation}
where $\lambda_{\text{CE}}, \lambda_{\text{BCE}}, \lambda_{\text{Dice}}$ are empirically determined weights for each term. 
Finally, the model is trained in an end-to-end manner using a total loss that combines both objectives as follows:
\begin{equation}
\mathcal{L} = \mathcal{L}_{\text{txt}} + \mathcal{L}_{\text{mask}}.
\label{eq:joint_loss}
\end{equation}

\begin{figure}[t]
    \centering
    \includegraphics[width=\columnwidth]{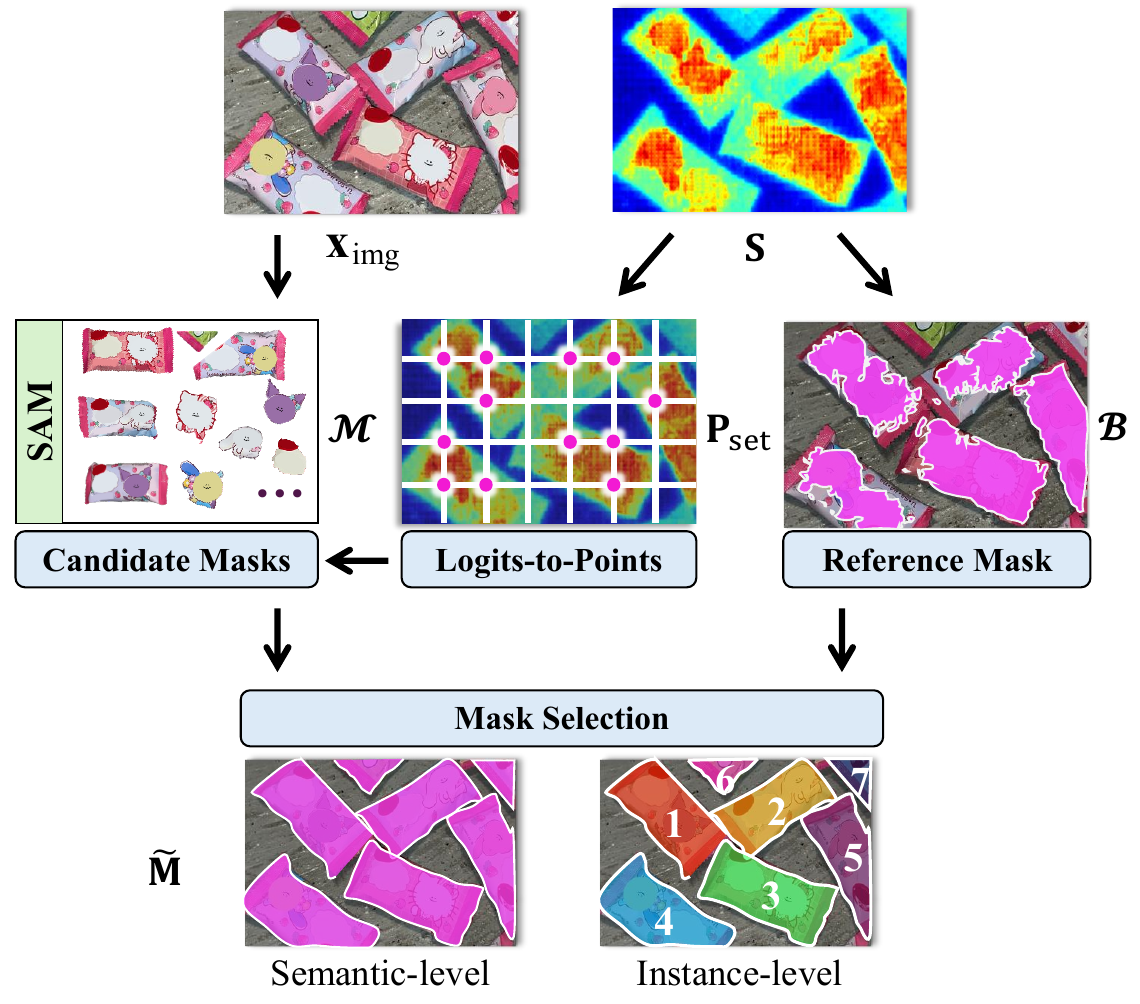}
    \caption{
    Overview of the NOVO Refinement.
    Without any additional training, our refinement method effectively combines the initial mask with SAM’s segmentation capability, not only enhancing the overall segmentation quality but also enabling instance-level reasoning segmentation.
    }
    \label{fig:sec_figure}
\end{figure}

\subsection{NOVO Refinement for High-Quality Instance-level Segmentation}
\label{sec:4.5}
While the proposed NOVO achieves high accuracy in reasoning segmentation, its predicted masks $\hat{\mathbf{M}}$ often exhibit visual artifacts such as holes and inaccurate boundaries.
These artifacts result from directly binarizing the segmentation logits $\mathbf{S}$ using a fixed threshold, a process which fails to capture the fine-grained structure of the target region.

To address this, we propose a refinement strategy that leverages the high-quality mask generation capability of the Segment Anything Model (SAM).
Rather than producing the final mask from binarized logits, we use $\mathbf{S}$ to generate point prompts $\mathbf{P}_{\text{set}}$, which guide SAM to produce precise instance-level mask candidates, resulting in a refined output mask $\widetilde{\mathbf{M}}$.
This strategy not only improves the visual quality of segmentation but also enables instance reasoning segmentation without additional training.  
The proposed refinement has three steps: (1) logits-to-point sampling, (2) reference mask construction, and (3) mask selection, as shown in Fig.~\ref{fig:sec_figure}.

First, to construct the point prompts $\mathbf{P}_{\text{set}}$, we sample locations from the segmentation logits $\mathbf{S} \in \mathbb{R}^{h \times w}$.   
We overlay a fixed-interval grid on $\mathbf{S}$ and select grid points whose values exceed $0$.  
The sampled points $\mathbf{P}_{\text{set}}$ are used as visual prompts to SAM, along with the input image $\mathbf{x}_{\text{img}}$.  
SAM then generates a set of $J$ instance-level candidate masks as follows:  
\begin{equation}
\mathcal{M} = \{ M_1, M_2, \ldots, M_J \}, \quad M_j \in \{0, 1\}^{h \times w}.
\end{equation} 
Each point in $\mathbf{P}_{\text{set}}$ acts as a prompt, leading to one or more candidate masks being generated in its vicinity.

Second, to guide the selection of masks from the candidate set $\mathcal{M}$, we define a binary reference mask $\mathcal{B}$ based on the distribution of the logits $\mathbf{S}$. 
We apply a fixed threshold $\tau=0$ to binarize the logits by $\mathcal{B} = \mathds{1}[\mathbf{S} \geq \tau]$.
% While our method adopts a fixed threshold for simplicity, alternative strategies—such as applying the Otsu algorithm~\cite{otsu1975threshold} to compute adaptive thresholds per image—could be explored in future work. 
The reference mask $\mathcal{B}$ serves as the criterion for evaluating and selecting high-quality masks from $\mathcal{M}$.
We then compute an overlap score between each candidate mask $M_j \in \mathcal{M}$ and the reference mask $\mathcal{B}$.
% to construct the refined output $\tilde{M}$.
The score measures how well each candidate $M_j$ aligns with the reference mask, and is defined as:
\begin{equation}
\mathrm{overlap}(M_j, \mathcal{B}) = \frac{|M_j \cap \mathcal{B}|}{|M_j|}.
\end{equation}

Finally, we select masks with scores exceeding a predefined threshold $\delta$ (empirically set to 0.7) to construct the refined output $\widetilde{\mathbf{M}}$.
Spatially overlapping masks are merged into a single output mask via union.
Importantly, our refinement strategy naturally supports both semantic and instance-level segmentation without additional model training:
\begin{itemize}
  \item \textbf{Semantic-level output}: All selected masks are aggregated to form a unified segmentation map with improved object completeness and boundaries.
  \item \textbf{Instance-level output}: Each selected mask without spatial overlap is treated independently for instance-wise output, enabling instance segmentation.
\end{itemize}

\section{\texttt{RISeg}: Dataset for Instance Reasoning Segmentation}
\label{sec:5}

% 인스턴스 수준의 reasoning segmentation 성능을 평가하기 위해, 새로운 벤치마크 데이터셋 \texttt{RISeg}을 제안한다.  
% \texttt{RISeg}은 COCO2017~\cite{lin2014microsoft} 검증 세트를 기반으로 구축되었으며, 총 918장의 이미지로 구성되어 있다.  
% 각 이미지에는 reasoning 중심의 텍스트 질의가 1개씩 포함되며, 하나 이상의 정답 인스턴스 마스크가 주석되어 총 2,533개의 인스턴스를 포함한다.  
% 이 인스턴스들은 COCO 카테고리 계층에 따라 12개의 상위 범주(supercategory)와 74개의 세분화된 객체 클래스(fine-grained object class)에 걸쳐 분포한다.
To evaluate the performance of instance-level reasoning segmentation, we introduce a new benchmark dataset, \texttt{RISeg}.
Built upon the COCO2017~\cite{lin2014microsoft} validation set, \texttt{RISeg} includes 918 images, each paired with a reasoning-focused text query and annotated with one or more ground-truth instance masks--totaling 2,533 instances.
% \tp{Unlike existing referring expression datasets such as RefCOCO, which primarily rely on explicit descriptions, \texttt{RISeg} incorporates more complex and implicit reasoning-oriented queries, requiring higher-level contextual understanding.}
Unlike \texttt{RefCOCO/+/g}~\cite{kazemzadeh2014referitgame, mao2016generation}, which uses explicit referring expressions, \texttt{RISeg} introduces implicit reasoning-based queries requiring deeper context.
% These instances span 12 supercategories and 74 fine-grained object classes, following the category structure used in COCO. An overview of \texttt{RISeg} is shown in Fig.~\ref{fig:dataset_figure}.
These instances are distributed across 6 supercategories--person, animal, food \& kitchenware, object \& furniture, transport \& outdoor, and others--and 74 fine-grained object classes.

\begin{figure}[t]
    \centering
    \includegraphics[width=\columnwidth]{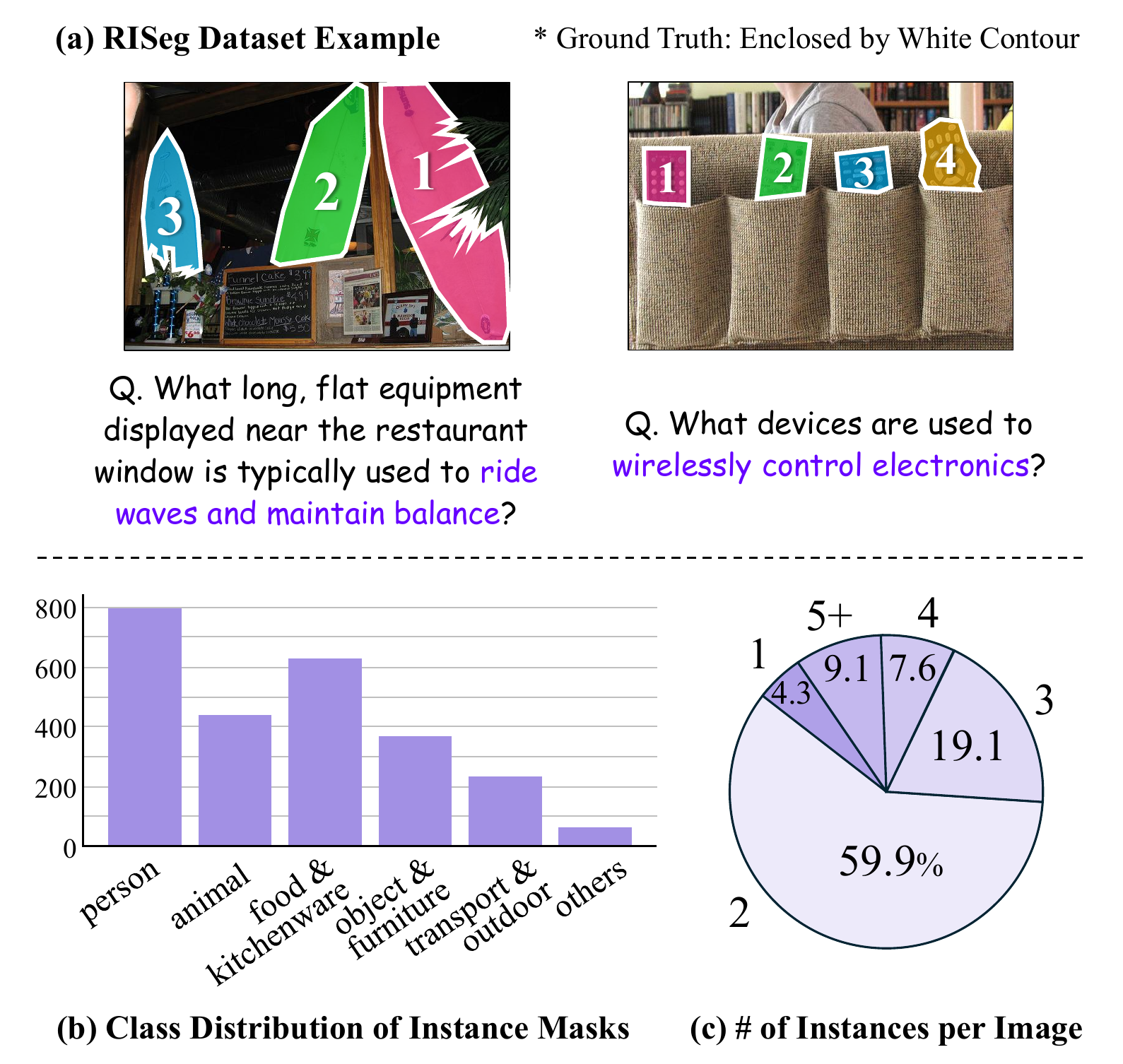}
    \caption{
    % \texttt{RISeg} 데이터셋 개요
    % (a) \texttt{RISeg}에서 사용된 이미지-질의-마스크 삼중쌍의 예시. 각 이미지는 추론 기반 텍스트 질의 및 해당 인스턴스 마스크와 쌍을 이룹니다. 흰색 윤곽선은 정답 마스크(ground-truth)를 나타냅니다.
    % (b) COCO의 상위 범주(supercategories) 기준으로 그룹화된 인스턴스 마스크 클래스 분포.
    % (c) 이미지당 주석된 인스턴스 수에 대한 히스토그램.
    % \texttt{RISeg}는 총 918개의 질의-이미지 쌍과 2,533개의 주석된 인스턴스를 포함하며, 이는 12개의 상위 범주에 걸쳐 분포되어 있습니다.
    Overview of the \texttt{RISeg} Dataset.
    (a) Examples from the \texttt{RISeg} dataset, where each image is paired with a reasoning-based text query and multiple ground-truth instance masks (shown with white contours).  
    % (b) Distribution of instance mask classes grouped by COCO supercategories. 
    (b) Distribution of instance mask classes grouped by categories.
    (c) The number of annotated instances per image.  
    % \texttt{RISeg} contains 918 reasoning queries over images, totaling 2,533 annotated instance masks spanning 6 supercategories.
    }
    \label{fig:dataset_figure}
\end{figure}

An overview of \texttt{RISeg} is shown in Fig.~\ref{fig:dataset_figure}.
% \texttt{RISeg} 데이터셋은 다음의 네 단계를 거쳐 구축되었다:  
% (1) 이미지 선택: 동일한 객체 클래스가 여러 개 포함된 이미지를 COCO 검증 세트에서 선별하였다.  
% (2) 질의 생성: 이미지, 캡션, 객체 라벨을 기반으로 대규모 언어 모델(GPT-4 등)을 활용하여 reasoning 기반 후보 질의를 생성하였다. 이때 질의는 객체의 속성, 공간적 관계, 맥락적 추론을 반영하도록 설계되었다.  
% (3) 사람 검증: 생성된 질의는 주석자가 검토하여 의미적 명확성과 논리적 타당성을 확보하였다.  
% (4) 마스크 정합: 마지막으로 검증된 질의를 해당 정답 인스턴스 마스크와 매칭하였다.  
% \texttt{RISeg}에 대한 보다 자세한 내용은 보충 자료에 포함되어 있다.
The construction of \texttt{RISeg} follows a four-step pipeline:
% (1) Image selection: Images containing multiple instances of the same object class are selected from the COCO validation set.
% (2) Query generation: Reasoning-based candidate queries are generated using a large language model (e.g., GPT-4), guided by the image, its caption, and object labels. The prompts are designed to reflect object attributes, spatial relations, and contextual reasoning.
(1) Image selection: Images containing multiple instances of the same class are selected from the COCO validation set.
(2) Query generation: Reasoning queries are generated using a large language model (e.g., GPT-4), guided by the image, its caption, and object labels. The prompts are designed to reflect object attributes, spatial relations, and contextual reasoning.
(3) Human validation: Each generated query is reviewed by human annotators to ensure semantic clarity and logical correctness.
(4) Mask alignment: Finally, validated queries are paired with their corresponding ground-truth instance masks.
Further details are provided in the supplementary materials.

\renewcommand{\arraystretch}{0.85}  % 전체적으로 촘촘한 표 유지
\begin{table*}[t]
\centering
\footnotesize
% \small
\resizebox{\textwidth}{!}{%
\begin{tabular}{l|cc|cc|cc|cc}
\specialrule{1.2pt}{0pt}{0pt}  % 굵은 선 (위 또는 아래에 사용)
\multirow{2}{*}{\textbf{Methods}} 
& \multicolumn{2}{c|}{val (overall)} & \multicolumn{2}{c|}{test (short query)} & \multicolumn{2}{c|}{test (long query)} & \multicolumn{2}{c}{test (overall)} \\
\cline{2-3} \cline{4-5} \cline{6-7} \cline{8-9}
& gIoU & cIoU & gIoU & cIoU & gIoU & cIoU & gIoU & cIoU  \\
% \specialrule{1.2pt}{0pt}{0pt}  % 굵은 선
\specialrule{0.4pt}{0pt}{0pt}  % 굵은 선
OVSeg {\scriptsize\cite{liang2023open}} & 28.5 & 18.6 & 18.0 & 15.5 & 28.7 & 22.5 & 26.1 & 20.8 \\
GRES {\scriptsize\cite{liu2023gres}} & 22.4 & 19.9 & 17.6 & 15.0 & 22.6 & 23.8 & 21.3 & 22.0 \\
X-Decoder {\scriptsize\cite{zou2023generalized}} & 22.6 & 17.9 & 20.4 & 11.6 & 22.2 & 17.5 & 21.7 & 16.3 \\
SEEM {\scriptsize\cite{zou2023segment}} & 25.5 & 21.2 & 20.1 & 11.5 & 25.6 & 20.8 & 24.3 & 18.7 \\
Grounded-SAM {\scriptsize\cite{liu2024grounding}} & 26.0 & 14.5 & 17.8 & 10.8 & 22.4 & 18.6 & 21.3 & 16.4 \\
\specialrule{0.4pt}{0pt}{0pt}
SESAME {\scriptsize\cite{wu2024see}} & 34.8 & 39.1 & 28.3 & 27.6 & 31.6 & 32.7 & 30.5 & 30.4 \\
LLaVA1.5-7B + OVSeg {\scriptsize\cite{liu2023visual, liang2023open}} & 38.2 & 23.5 & 24.2 & 18.7 & 44.6 & 37.1 & 39.7 & 31.8 \\
LISA-7B (ft) {\scriptsize\cite{lai2024lisa}} & 52.9 & 54.0 & 40.6 & 40.6 & 49.4 & 51.0 & 47.3 & 48.4 \\
LISA-7B-LLaVA1.5 (ft) {\scriptsize\cite{lai2024lisa}} & 61.3 & 62.9 & 48.3 & 46.3 & 57.9 & 59.7 & 55.6 & 56.9 \\
LISA++-7B-LLaVA1.5 (ft) {\scriptsize\cite{yang2023lisa++}} & 64.2 & 68.1 & 49.6 & \textbf{51.1} & 59.3 & 61.7 & 57.0 & 59.5 \\
READ-7B-LLaVA1.5 (ft) {\scriptsize\cite{qian2025reasoning}} & 59.8 & 67.6 & 52.6 & 49.5 & 60.4 & 61.0 & 58.5 & 58.6 \\
RSVP-GPT-4o {\scriptsize\cite{lu2025rsvp}} & 64.7 & 63.1 & \textbf{55.4} & 50.4 & 61.9 & 62.5 & 60.3 & 60.0 \\
\textbf{Ours-7B-LLaVA1.5 (ft)} & \textbf{66.3} & \textbf{69.0} & 53.1 & 49.5 & \textbf{65.4} & \textbf{70.3} & \textbf{62.4} & \textbf{65.7} \\
\specialrule{0.4pt}{0pt}{0pt}
LLaVA1.5-13B + OVSeg {\scriptsize\cite{liu2023visual, liang2023open}} & 37.9 & 26.4 & 27.1 & 19.4 & 46.1 & 40.6 & 41.5 & 34.1 \\
LISA-13B-LLaVA1.5 (ft) {\scriptsize\cite{lai2024lisa}} & 65.0 & \textbf{72.9} & 55.4 & 50.6 & 63.2 & 65.3 & 61.3 & 62.2 \\
READ-13B-LLaVA1.5 (ft) {\scriptsize\cite{qian2025reasoning}} & - & - & 55.4 & 53.7 & 64.4 & 65.1 & 62.2 & 62.8 \\
\textbf{Ours-13B-LLaVA1.5 (ft)} & \textbf{66.7} & 71.9 & \textbf{57.5} & \textbf{57.0} & \textbf{67.8} & \textbf{68.1} & \textbf{65.3} & \textbf{66.0} \\
\specialrule{1.2pt}{0pt}{0pt}
\end{tabular}
}
\caption{Reasoning segmentation performance on the \texttt{ReasonSeg} dataset. (ft) denotes a model fine-tuned on 239 samples from the ReasonSeg training set.
Bold numbers indicate the best performance among 7B-based and 13B-based models, respectively.}
\label{tab:reasonseg_full}
\end{table*}

\begin{table}[tb]
\centering
% \footnotesize
% \small
\renewcommand{\arraystretch}{1}
\resizebox{\linewidth}{!}{%
\begin{tabular}{l|ccc|ccc}
\specialrule{1.2pt}{0pt}{0pt}  % 굵은 선 (위 또는 아래에 사용)
% \textbf{Method} 
\multirow{2}{*}{\textbf{Methods}} 
& \multicolumn{3}{c|}{val (overall)} & \multicolumn{3}{c}{test (overall)} \\
\cline{2-4} \cline{5-7}
\rule{0pt}{1.6ex} & gIoU & B-IoU & B-F1 & gIoU & B-IoU & B-F1 \\
\specialrule{1.2pt}{0pt}{0pt}  % 굵은 선 (위 또는 아래에 사용)
READ-7B$^\dagger$ & 59.8 & 29.1 & 38.9 & 56.3 & 30.4 & 39.8 \\
READ-7B (+R)$^\dagger$ & \textbf{61.3} & \textbf{36.1} & \textbf{46.1} & \textbf{56.4} & \textbf{34.6} & \textbf{44.3} \\
% READ-7B (+R adaptive) & 61.3 & 36.0 & 46.0 & - & - & - \\
\hline
Ours-7B & 66.3 & 33.9 & 44.0 & 62.4 & 33.6 & 42.8 \\
Ours-7B (+R) & \textbf{67.1} & \textbf{40.0} & \textbf{50.3} & \textbf{62.7} & \textbf{39.1} & \textbf{48.6} \\
% Ours-7B (+R adaptive)\rule{0pt}{1.5ex} & \textbf{67.6} & \textbf{41.0} & \textbf{51.5} & \textbf{62.7} & \textbf{39.2} & \textbf{48.7} \\
\hline
Ours-13B & 66.7 & 35.5 & 45.1 & \textbf{65.3} & 36.5 & 46.0 \\
Ours-13B (+R) & \textbf{67.2} & \textbf{40.9} & \textbf{51.3} & 65.2 & \textbf{40.6} & \textbf{50.2} \\
% Ours-13B (+R adaptive)\rule{0pt}{1.5ex} & \textbf{67.2} & 40.8 & 51.0 & \textbf{65.6} & \textbf{40.9} & \textbf{50.6} \\
\specialrule{1.2pt}{0pt}{0pt}  % 굵은 선 (위 또는 아래에 사용)
\end{tabular}%
}
\caption{
% \texttt{ReasonSeg} 데이터셋에서 RSVP 정제 모듈(Sec.~4.5) 적용 전후의 시맨틱 분할 성능.  
% 정제 모델(+R)은 검증 및 테스트 셋 모두에서 gIoU, 경계 IoU(B-IoU), 경계 F1(B-F1)을 향상시킨다.  
% 모든 경계 지표는 3픽셀 허용 오차로 계산된다.
Semantic segmentation results on \texttt{ReasonSeg} with and without NOVO Refinement. (+R) denotes our refinement. $^\dagger$ indicates values reproduced in our experimental setting.
%(3-pixel boundary tolerance). 
% The refined models (+R) improve gIoU, Boundary IoU, and Boundary F1 on both validation and test sets.
}
\label{tab:refinement}
\end{table}

\section{Experimental Results}
\label{sec:6}
\subsection{Settings}
\label{sec:6.1}
\noindent \textbf{Implementation Details.} \quad
In this study, we adopt LLaVA 1.5-7B and LLaVA 1.5-13B~\cite{liu2023visual} as the Vision-Language Model (VLM) $\mathcal{F}_{\text{VL}}$, and employ the Segment Anything Model (SAM)~\cite{kirillov2023segment} with a ViT-H backbone as the segmentation module $\mathcal{F}_{\text{S}}$.
All experiments were conducted using the DeepSpeed framework on four NVIDIA A100 GPUs (40GB each), with a total training time of approximately 26 hours. 
Input images were padded to square shapes before being fed into the network, and outputs were restored to their original aspect ratios for evaluation.
We used the AdamW optimizer with a learning rate of 0.0005 and a learning rate scheduler with 100 warm-up steps. 
The loss weights were set to $\lambda_{\text{CE}} = 1.0$ for text generation, and $\lambda_{\text{BCE}} = 2.0$ and $\lambda_{\text{Dice}} = 0.5$ for mask prediction.
In NOVO, we apply LoRA~\cite{hu2022lora} to the VLM for parameter-efficient tuning, and jointly train the mask decoder in SAM's segmentation module. All other components, including the image encoder and prompt encoder in SAM, as well as the remaining parts of the VLM, are frozen.

\begin{table}[tb]
\centering
% \footnotesize
\renewcommand{\arraystretch}{1.1}
\resizebox{\linewidth}{!}{%
\begin{tabular}{l|ccc|ccc}
\specialrule{1.2pt}{0pt}{0pt}  % 굵은 선 (위 또는 아래에 사용)
\textbf{Methods} & AP50 & AP75 & mAP & AP-S & AP-M & AP-L \\
\specialrule{1.2pt}{0pt}{0pt}  % 굵은 선 (위 또는 아래에 사용)
LISA++-7B & 14.6 & 5.0 & 6.1 & 1.1 & 2.5 & 10.6 \\
% LISA (+R) & - & - & - & - & - & - \\
\hline
READ-7B (+R) & 43.2 & \textbf{30.1} & 27.2 & 7.8 & 18.7 & \textbf{38.6} \\
Ours-7B (+R) & \textbf{44.7} & 30.0 & \textbf{27.2 }& \textbf{9.8} & \textbf{19.9} & 37.0 \\
\specialrule{1.2pt}{0pt}{0pt}  % 굵은 선 (위 또는 아래에 사용)
\end{tabular}
}
\caption{
% RSVP 정제 모듈(Sec.~4.5)을 적용한 \texttt{RISeg} 벤치마크의 인스턴스 분할 성능.  
% \(\mathcal{B}_{\text{M}}\)을 참조 마스크로 사용한 정제 모델(+R)은 모든 AP 지표 및 객체 크기에서 LISA++보다 우수하다.
Instance-level reasoning segmentation results on \texttt{RISeg} dataset. (+R) denotes our Refinement method.
% The refined models (+R) outperform LISA++ across all AP metrics and object sizes.
}
\label{tab:our_dataset_instance}
\end{table}

\begin{figure*}[t]
    \centering
    % \hspace*{0.15\textwidth}% <-- 왼쪽에 15%의 여백 추가    
    \includegraphics[width=\textwidth]{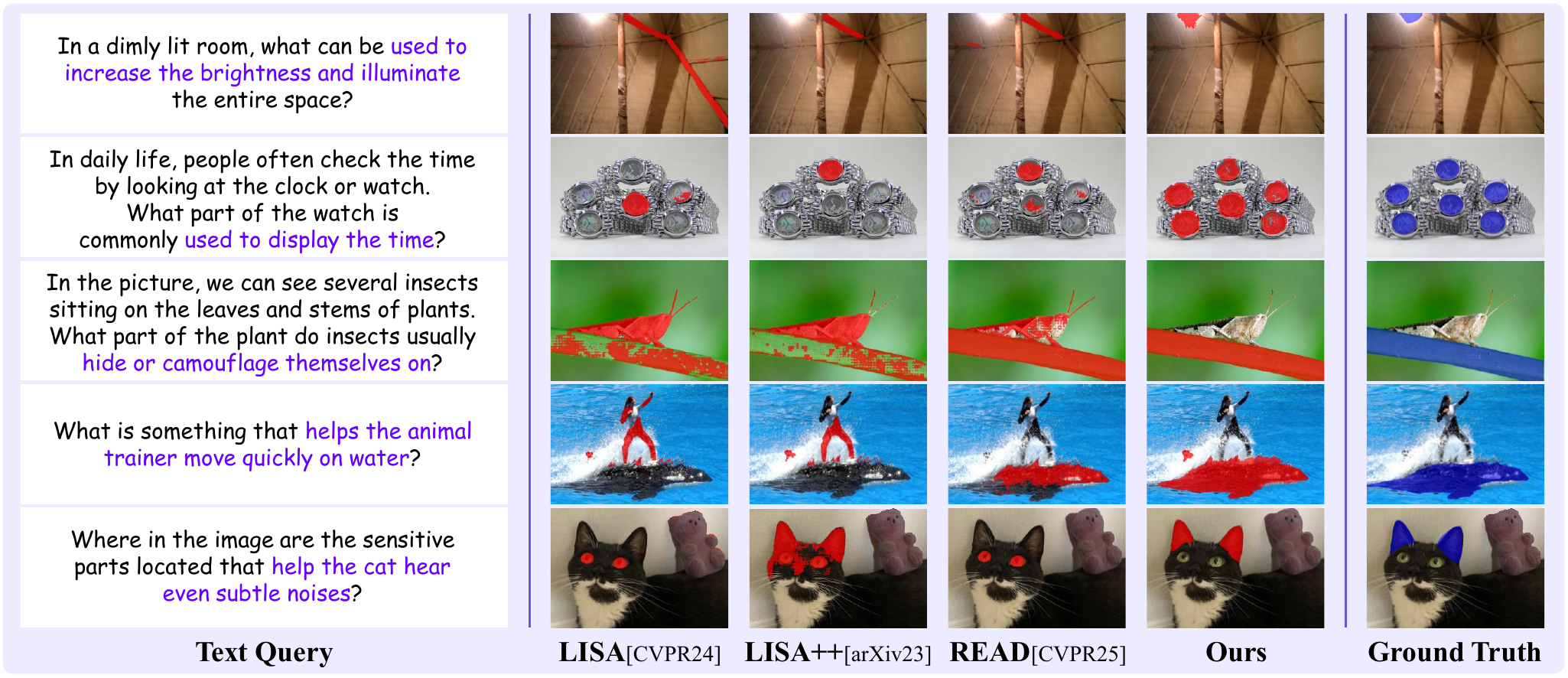}
    \caption{
    Qualitative comparison of NOVO and existing methods on reasoning segmentation tasks. 
    NOVO produces accurate and coherent masks, even in challenging reasoning cases such as ambiguous boundaries and multi-instance scenarios. More qualitative examples are provided in the supplementary materials.}
    \label{fig:quantitative_result}
\end{figure*}

\noindent \textbf{Datasets.} \quad
We use a range of datasets relevant to reasoning segmentation for training and evaluation:
{(1) Semantic segmentation}: \texttt{ADE20K}~\cite{zhou2017scene}, \texttt{COCO-Stuff}~\cite{caesar2018coco}, \texttt{PACO-LVIS}~\cite{ramanathan2023paco}, and \texttt{PASCAL-Part}~\cite{chen2014detect};
{(2) Referring segmentation}: \texttt{refCLEF}, \texttt{RefCOCO}, \texttt{RefCOCO+}~\cite{kazemzadeh2014referitgame}, \texttt{RefCOCOg}~\cite{mao2016generation}, and \texttt{ReasonSeg}~\cite{lai2024lisa};
{(3) Visual question answering}: \texttt{LLaVA Instruct 150K}~\cite{liu2023visual}.
To further enhance the model’s ability to handle false premises and reasoning mismatches, we additionally use \texttt{FP-RefCOCO(+/g)}~\cite{wu2024see} and \texttt{R-RefCOCO}~\cite{wu2024toward}.

\noindent \textbf{Evaluation Metrics.} \quad
We evaluate segmentation performance based on two main criteria: semantic-level and instance-level accuracy.  
For semantic-level evaluation, we follow prior works~\cite{lai2024lisa, qian2025reasoning} and evaluate both global IoU (gIoU) and cumulative IoU (cIoU).
The gIoU is computed as the average IoU across individual images, while the cIoU is calculated as the intersection over union aggregated over the entire dataset.
Since cIoU is more sensitive to object size and may introduce bias, we focus on gIoU as it provides a more balanced evaluation across different object sizes.
In addition, to assess the boundary quality of the refined masks introduced in Sec.\ref{sec:4.5}, we employ Boundary IoU and Boundary F1 metrics~\cite{cheng2021boundary}.
For instance-level evaluation, we measure the mean Average Precision (mAP), following~\cite{yang2023lisa++}.

\begin{figure}[t]
    \centering
    \includegraphics[width=\columnwidth]{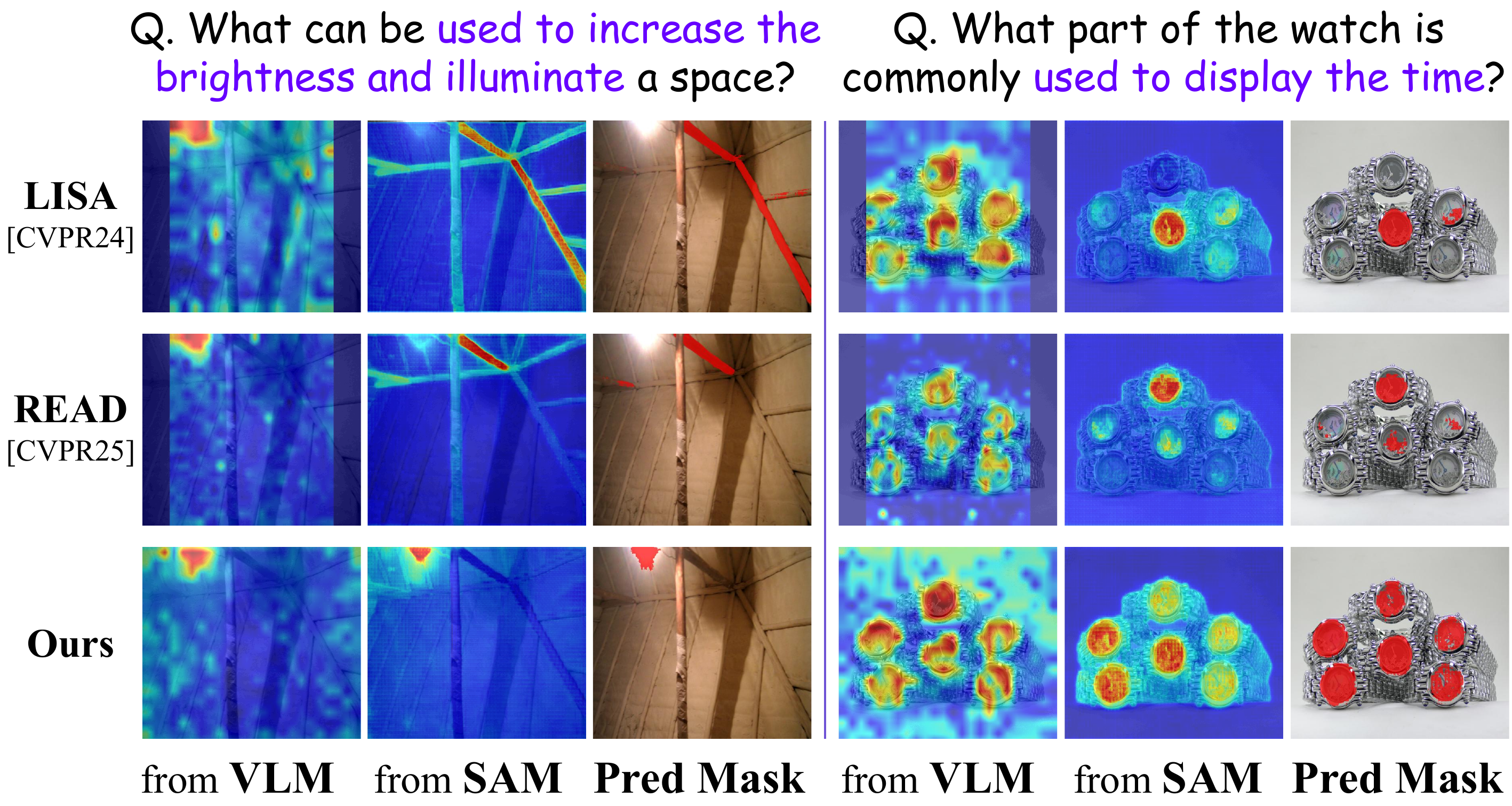}
    \caption{
    % VLM activation map, SAM 마스크 로짓, 최종 예측 마스크 간 모델 비교.
    % LISA와 READ는 VLM과 SAM 사이의 정합이 부족한 반면, RSVP는 두 단계 간 일관성을 유지하여 보다 정확하고 의미적으로 일치하는 마스크를 생성한다. (Figure~\ref{fig:quantitative_result}의 1행과 2행 사례 사용)
    Comparison of VLM activation maps, SAM's mask logits, and predicted masks across different models.
    }
    \label{fig:qulitive_figure3}
\end{figure}

\subsection{Performance Comparison}
\label{sec:6.2}
Table~\ref{tab:reasonseg_full} compares the reasoning segmentation performance of our proposed NOVO model with existing state-of-the-art methods.
7B and 13B refer to the VLM baseline models LLaVA 1.5-7B and LLaVA 1.5-13B, respectively.
Ours-7B achieved 66.3\% gIoU and 69.0\% cIoU on the validation set, outperforming READ-7B by +6.5\% and +1.4\%, respectively.
On the test set, it recorded 62.4\% gIoU and 65.7\% cIoU, showing improvements of +3.9\% and +7.1\% over READ-7B, while delivering performance comparable to larger 13B-scale models.
Furthermore, Ours-13B achieved the highest performance among all methods, with 66.7\% gIoU and 71.9\% cIoU on the validation set, and 65.3\% gIoU and 66.0\% cIoU on the test set.
These results highlight the effectiveness of our approach. Specifically, our method converts VLM outputs into visual prompts, allowing the segmentation model to generate more accurate and robust masks.

\begin{figure}[t]
    \centering
    \includegraphics[width=\columnwidth]{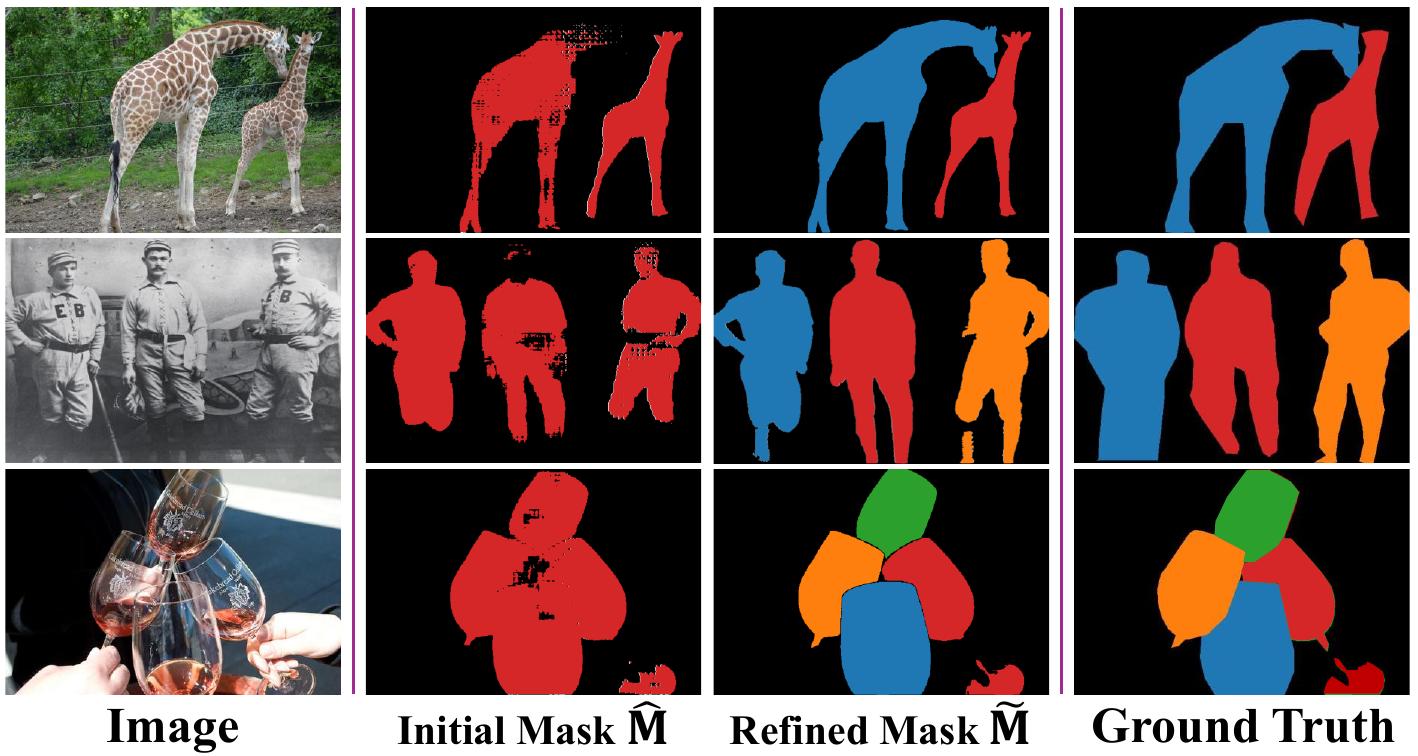}
    \caption{
    Qualitative comparison of reasoning segmentation results. Different colors indicate different object instances.}
    % The refined masks $\tilde{M}$ show reduced artifacts and more accurate instance boundaries compared to the binarized masks $\hat{M}$.
    \label{fig:sec_figure2}
\end{figure}
%(yjcho)
% VLM, SAM logit의 개선을 보여줄수 있는 실험 결과 및 해석 추가.
% \tb{Figure~\ref{fig:quantitative_result} presents a qualitative comparison with prior models, highlighting the effectiveness of our method.
% The comparison includes LISA-7B, LISA++-7B-LLaVA1.5, READ-7B-LLaVA1.5, and Ours-7B-LLaVA1.5.
% Our RSVP consistently generates more accurate and semantically coherent masks in response to complex reasoning queries.
% In particular, our method performed well in challenging cases, such as detecting objects with ambiguous boundaries (e.g., row 1) and identifying multiple instances simultaneously (e.g., row 2).
% Additional qualitative examples are provided in the supplementary materials.}
% Figure~4, 5는 기존 모델들과의 정성적 비교를 제시하며, 본 방법의 효과를 강조한다. 비교 대상에는 LISA-7B, LISA++-7B-LLaVA1.5, READ-7B-LLaVA1.5, Ours-7B-LLaVA1.5가 포함된다. 
% 그림~\ref{fig:vlm_sam_mismatch}은 LISA, READ, 그리고 제안한 방법에 대해 VLM의 coarse activation map, SAM의 마스크 로짓(SAM’s mask logits), 그리고 최종 예측 마스크를 비교한 것이다.  
% LISA와 READ는 VLM 수준에서는 질의와 관련된 영역을 어느 정도 잘 포착하지만, 이후 단계인 SAM에서는 비관련 영역에 주목하여 부정확한 마스크를 생성하는 경향이 있다.  
% 반면, 제안한 방법은 VLM의 coarse activation map과 SAM 출력 간의 일관성을 유지하며, 더 정확하고 의미적으로 일치하는 마스크를 생성한다.  
% 이는 사전 학습된 SAM의 segmentation 능력을 reasoning 기반 질의에 효과적으로 연계하도록 설계된 본 방법의 유효성을 보여준다.
%Figures~\ref{fig:quantitative_result} and \ref{fig:qulitive_figure3} present a qualitative comparison with prior models, highlighting the effectiveness of our method.  
%The comparison includes LISA-7B, LISA++-7B-LLaVA1.5, READ-7B-LLaVA1.5, and Ours-7B-LLaVA1.5.
Figure~\ref{fig:quantitative_result} provides qualitative results of the NOVO model on complex reasoning segmentation tasks.  
Our model performs well in challenging cases, such as detecting objects with ambiguous boundaries (1\textsuperscript{st} row) and identifying multiple instances simultaneously (2\textsuperscript{nd} row).
These results validate the effectiveness of NOVO in consistently generating query-aligned masks across diverse and complex reasoning scenarios.
Additional qualitative examples are provided in the supplementary materials.
Figure~\ref{fig:qulitive_figure3} presents a visual comparison of three stages: the coarse activation maps from the VLM, the mask logits from SAM, and the final predicted masks. We show results for LISA~\cite{lai2024lisa}, READ~\cite{qian2025reasoning}, and our proposed method at each stage. 
While LISA and READ reasonably highlight relevant regions in the VLM activation maps, they often produce inaccurate masks at the SAM stage. 
In contrast, our method maintains consistency between the VLM's coarse activations and SAM's outputs, leading to more precise and semantically coherent masks.
This demonstrates the effectiveness of our design in leveraging SAM’s pretrained segmentation capabilities.

% 그림~\ref{fig:quantitative_result}은 복잡한 reasoning segmentation 과제에 대한 RSVP 모델의 정성적 성능을 추가로 보여준다.  
% 제안한 모델은 경계가 모호한 객체를 식별해야 하는 경우(예: 1행)나 여러 인스턴스를 동시에 탐지해야 하는 경우(예: 2행)에서도 뛰어난 성능을 보인다.  
% 이는 RSVP가 다양한 reasoning 상황에서도 텍스트 질의에 부합하는 마스크를 안정적으로 생성함을 보여준다.  
% 추가적인 정성적 예시는 부록에 수록되어 있다.
% 또한 \textbf{Ours-13B-LLaVA1.5 (ft)} 모델은 validation set에서 gIoU 66.7\%, cIoU 71.9\%, test set에서 gIoU 65.3\%, cIoU 66.0\%를 기록하며, 모든 비교 방법 중 최고 성능을 보였다.
%Furthermore, Ours-13B model achieved the best performance among all methods, with 66.7\% gIoU and 71.9\% cIoU on the validation set, and 65.3\% gIoU and 66.0\% cIoU on the test set.
% 이러한 결과는 VLM의 출력을 visual prompt로 변환하여 segmentation 모델에 전달하는 본 방법이, 다양한 reasoning query에 대해 더욱 정밀하고 강건한 마스크를 생성함을 시사한다.
%These results highlight the strength of our approach, where the VLM outputs are converted into visual prompts, enabling the segmentation model to produce more precise and robust masks for complex reasoning queries.

% 6.4 Instance-level Reasoning Segmentation Results
\subsection{Effectiveness of NOVO Refinement}
\label{sec:6.3}
% RSVP Refinement의 효과를 평가하기 위해, 본 절에서는 시맨틱 수준 및 인스턴스 수준 분할 평가를 모두 수행한다.  
% 정제 성능을 보다 정밀하게 평가하기 위해, gIoU, Boundary IoU (B-IoU), 그리고 Boundary F1 (B-F1)를 평가 지표로 사용한다.
In this section, we assess the effectiveness of NOVO Refinement through both semantic-level and instance-level segmentation evaluations.
% To better assess the performance of the refinement, we measure gIoU, Boundary IoU (B-IoU), and Boundary F1 (B-F1) as evaluation metrics.

% Table~\ref{tab:refinement}은 \texttt{ReasonSeg} 데이터셋에서의 시맨틱 분할 성능을 비교한 것이다.  
% 모든 경계 기반 평가는 3픽셀의 허용 오차를 기준으로 계산된다.  
% 정제 모델(+R)은 validation 및 test 세트 모두에서 gIoU, Boundary-IoU, Boundary-F1 점수를 꾸준히 향상시킨다.  
% 이 결과는 제안된 정제 모듈이 특히 객체의 경계 주변에서 마스크 품질을 효과적으로 향상시킴을 보여준다.
Table~\ref{tab:refinement} compares semantic-level segmentation performance on the \texttt{ReasonSeg} dataset.  Boundary-related metrics such as Boundary-IoU and Boundary-F1 were computed with a 3-pixel tolerance.
On both the validation and test sets, our refined models (+R) achieved consistent improvements in gIoU, B-IoU, and B-F1.
These results demonstrate that our refinement module effectively improves mask quality, particularly around object boundaries.
% Table~\ref{tab:our_dataset_instance}는 \texttt{RISeg} 벤치마크에서의 인스턴스 수준 분할 성능을 보여준다.  
% 정제 모델(+R)은 모든 AP 지표와 객체 크기(AP-S, AP-M, AP-L)에서 기존 방법인 LISA++보다 뛰어난 성능을 보인다.  
% 주목할 점은, 본래 시맨틱 분할을 위해 설계된 모델이 정제 전략을 통해 고품질의 인스턴스 마스크도 생성할 수 있다는 것이다.
% Table~\ref{tab:our_dataset_instance} presents the instance-level segmentation results on the \texttt{RISeg} benchmark.  
% The refined models (+R) outperformed the previous method LISA++~\cite{yang2023lisa++} across all AP metrics and object sizes.
Our NOVO Refinement also enables instance-level reasoning segmentation. 
To evaluate this capability, we conducted experiments on the \texttt{RISeg} dataset, as summarized in Tab.~\ref{tab:our_dataset_instance}. 
We compared our method with LISA++~\cite{yang2023lisa++}, a prior approach capable of instance-level reasoning. 
Regardless of the underlying baseline, our refined models (+R) consistently outperformed LISA++ across all AP metrics and object sizes.

% Figure~\ref{fig:sec_figure2}는 RSVP refinement 전후의 인스턴스 수준 reasoning segmentation 결과에 대한 정성적 비교를 보여준다.  
% 단순 임계값을 적용하여 얻은 초기 마스크 \(\hat{M}\)은 종종 구멍이나 부정확한 경계선과 같은 시각적 artifact를 포함한다.  
% 반면, 정제된 마스크 \(\tilde{M}\)은 적응형 마스크 생성을 통해 이러한 artifact를 효과적으로 완화하며 더욱 정밀한 분할 결과를 제공한다.  
% 또한, 제안된 RSVP refinement는 SAM의 분할 출력을 활용함으로써, 추가 학습 없이도 자연스럽게 인스턴스 수준 분할을 가능하게 한다.
Figure~\ref{fig:sec_figure2} presents a qualitative comparison of instance-level reasoning segmentation results before and after NOVO refinement.
The initial mask $\hat{\mathbf{M}}$, obtained by simple thresholding the segmentation logits, often contained visual artifacts such as holes and inaccurate boundaries.
In contrast, the refined masks $\widetilde{\mathbf{M}}$ effectively mitigate these artifacts through adaptive mask generation, resulting in more precise segmentation.
Moreover, the proposed NOVO refinement leverages SAM’s segmentation outputs, enabling instance-level segmentation as a natural outcome of the refinement process.

\subsection{Ablation Study on Prompt Types}
% 본 절에서는 서로 다른 프롬프트 유형의 효과를 분석하기 위해 \texttt{ReasonSeg} 검증 세트에서 gIoU와 cIoU를 활용한 ablation study를 수행하였다. 
% 모든 실험은 사전 학습된 LLaVA 기반 READ 모델~\cite{qian2025reasoning}을 비전-언어 백본으로 사용하였으며, 동일한 하이퍼파라미터 설정하에 학습을 진행하였다.
% 실험 결과는 Table~\ref{tab:sasp_ablation}에 요약되어 있다. 
% 단일 유형의 프롬프트를 사용할 경우, 마스크 프롬프트가 포인트 프롬프트보다 우수한 성능을 보였으며 (59.9 vs. 55.8 gIoU), 이는 마스크 단위의 단서가 세분화 모델에 더 강력한 공간적 지침을 제공함을 시사한다. 
% 그럼에도 불구하고 포인트 프롬프트 역시 주요 영역을 강조하는 데 효과적이었다. 
% 마스크와 포인트 프롬프트를 함께 사용할 경우 가장 높은 성능을 기록하였으며 (66.3 gIoU, 69.0 cIoU), 이는 포인트 프롬프트만 사용하는 설정 대비 각각 +6.4 및 +16.7 포인트 향상된 결과이다. 
% 이러한 결과는 마스크 프롬프트가 정밀한 위치 정보를 제공하는 반면, 포인트 프롬프트는 주요 영역을 강조하며, 두 방식이 상호 보완적으로 작용하여 reasoning segmentation 성능을 향상시킴을 보여준다.
In this section, we investigate the effect of different prompt types through an ablation study on the \texttt{ReasonSeg} validation set, using gIoU and cIoU as evaluation metrics. 
All experiments adopt the pretrained LLaVA-based READ model~\cite{qian2025reasoning} as the vision-language backbone and are trained under the same hyperparameters as our baseline.

The results are summarized in Table~\ref{tab:sasp_ablation}. 
When a single type of prompt is used, mask prompts outperform point prompts (59.9 vs. 55.8 gIoU), suggesting that mask-level cues provide stronger spatial guidance to the segmentation model. 
Nevertheless, point prompts remain effective in highlighting key regions. 
The combination of mask and point prompts yields the best performance overall (66.3 gIoU and 69.0 cIoU), representing relative improvements of +10.5 and +16.7 points compared to the point-only setting. 
This demonstrates that mask prompts deliver fine-grained localization, while point prompts emphasize salient regions. Together, they play complementary roles in enhancing reasoning segmentation.

\begin{table}[t]
\centering
\renewcommand{\arraystretch}{1}
\begin{tabular}{c|c|c|c|c}
\specialrule{1.2pt}{0pt}{0pt}  % 굵은 선 (위 또는 아래에 사용)
Exp. ID & $\mathbf{P}_{\text{mask}}$ & $\mathbf{P}_{\text{point}}$ & gIoU & cIoU \\
\hline
1 & \checkmark &            & 59.9 & 66.1 \\ %\hline
2 &            & \checkmark & 55.8 & 52.3 \\ %\hline
3 & \checkmark & \checkmark & \textbf{66.3} & \textbf{69.0} \\ %\hline
\specialrule{1.2pt}{0pt}{0pt}  % 굵은 선 (위 또는 아래에 사용)
\end{tabular}
\caption{
Ablation study on the use of mask and point prompts.
Combined prompts yield the best scores in both gIoU and cIoU.
}
\label{tab:sasp_ablation}
\end{table}

%%%%%%%%%%%%%%%%%%%%%%%%%%%%%%%%%%%%%%%%%%%%%%%%%%%%%%%%%%%%%%%%%%%%%%%%%%%%%%%%%%%%%%%%%%%%%%%%%
                                        % Conclusion
%%%%%%%%%%%%%%%%%%%%%%%%%%%%%%%%%%%%%%%%%%%%%%%%%%%%%%%%%%%%%%%%%%%%%%%%%%%%%%%%%%%%%%%%%%%%%%%%%

\section{Conclusions and Future Work}
% 우리는 RSVP라는 새로운 reasoning segmentation 패러다임을 제안했다. 이 프레임워크는 텍스트 기반이 아닌 시각 프롬프트만을 활용하여, 비전-언어 모델의 이해를 사전 학습된 분할 모델과 효과적으로 연결한다.
% 특히 VLM에서 coarse mask와 point prompt를 생성하여 SAM의 입력 형식에 맞추었고, 이를 통해 의미론적 불일치를 최소화하며 정확한 세분화를 가능케 했다.
% 추가 학습 없이 작동하는 refinement 모듈은 마스크 품질을 향상시킬 뿐만 아니라 인스턴스 수준 분할까지 자연스럽게 확장된다.
% RSVP는 ReasonSeg와 RISeg 벤치마크 모두에서 새로운 성능 기준을 제시하며, 시각 프롬프트 기반 분할의 가능성을 입증하였다.
In this study, we proposed NOVO, a new framework for reasoning segmentation that focuses solely on visual prompts, effectively bridging the gap between vision-language understanding and the visual prompt–based segmentation model SAM.
NOVO generates a coarse mask and point prompts compatible with SAM’s input format, avoiding semantic misalignment and enabling accurate segmentation.
A refinement module further improves mask quality and naturally extends NOVO to instance-level segmentation, all without additional training.
NOVO achieves new state-of-the-art results on both the \texttt{ReasonSeg} and \texttt{RISeg} benchmarks, demonstrating the effectiveness of visual prompt transformation for reasoning segmentation.

While NOVO demonstrates strong performance, it also has several limitations.
First, NOVO Refinement operates with fixed thresholds and heuristics. Although effective in removing boundary noise and improving mask sharpness, it may struggle with complex errors. A learning-based refinement could offer a more adaptive solution.
Second, our benchmark dataset \texttt{RISeg} is relatively small (918 images, 2,533 instance masks). While suitable for initial evaluation, its limited coverage of categories and reasoning types may restrict generalization. Future work could expand the dataset across more domains and reasoning dimensions.

As future work, we aim to extend NOVO to multi-label reasoning segmentation, where multiple relevant objects must be segmented and ranked according to a given query.
For example, the query ``Find the fruits with the highest vitamin content in order'' requires both object identification and semantic ranking.
This extension could further enhance the reasoning capability and applicability of NOVO.

{
    \small
    \bibliographystyle{ieeenat_fullname}
    \bibliography{main}
}

\end{document}